\documentclass{article} % For LaTeX2e
\usepackage{iclr2026_conference,times}
\usepackage{amsthm} 
\usepackage{enumitem}
\setlist[itemize]{itemsep=1pt, parsep=0pt, topsep=0pt, partopsep=0pt}
\usepackage{hyperref}
\usepackage{url}
\usepackage{comment}
\usepackage{algorithmic}
\usepackage{subfigure}
\usepackage{graphicx}
\usepackage{subcaption}
\usepackage{subfloat}
\usepackage{booktabs}
\usepackage{arydshln}
\usepackage{array}
\usepackage{multirow}
\usepackage{xspace}
\usepackage[symbol]{footmisc}
\title{
Latent Personality Alignment: Improving Harmlessness Without Mentioning Harms
}

\author{\textbf{Linh Le\textsuperscript{1,2}, 
    David Williams-King\textsuperscript{3},
    Mohamed Amine Merzouk\textsuperscript{1,2},}\\
    \textbf{Aton Kamanda\textsuperscript{4},
    Adam Oberman\textsuperscript{1,2,4}}
    \\
    \textsuperscript{1}McGill University,
    \textsuperscript{2}Mila – Quebec AI Institute,
    \textsuperscript{3}ERA,
    \textsuperscript{4}LawZero
    }

\newcommand{\pname}{LPA\xspace}
\newcommand{\pnameloop}{LPA-overfit\xspace}
\newcommand{\pnameflip}{LPA-flip\xspace}

\iclrfinalcopy % Uncomment for camera-ready version, but NOT for submission.
\begin{document}

\maketitle

%\vspace{-1em}
%\begin{center}
%\small\texttt{\{thai.linh.le, mohamed.merzouk\}@mila.quebec\quad snowelf42@gmail.com\quad atonkamanda@hotmail.com\quad adam.oberman@mcgill.ca}
%\end{center}

\begin{abstract}
Current adversarial robustness methods for large language models require extensive datasets of harmful prompts (thousands to hundreds of thousands of examples), yet remain vulnerable to novel attack vectors and distributional shifts. We propose Latent Personality Alignment (\pname), a sample-efficient defense that achieves robustness by training models on abstract personality traits rather than specific harmful behaviors. Using fewer than 100 trait statements and latent adversarial training, \pname achieves comparable attack success rates to methods trained on 150k+ examples, while maintaining superior utility. Critically, \pname generalizes better to unseen attack distributions, reducing misclassification rates by 2.6$\times$ compared to baseline across six harm benchmarks---without ever seeing harmful examples during training. Our results demonstrate that personality-based alignment offers a principled approach to building robust defenses with minimal cost.
\end{abstract}
\footnotetext[0]{Corresponding authors: 
        \texttt{\{thai.linh.le, mohamed-amine.merzouk\}@mila.quebec}; \texttt{david@erafellowship.org}; 
        \texttt{a.kamanda@lawzero.org}; 
        \texttt{adam.oberman@mcgill.ca}}

\section{Introduction}

Adversarial robustness in large language models (LLMs) presents a fundamental challenge: models must resist diverse attack vectors while maintaining utility on benign tasks. Current defenses typically rely on safety fine-tuning with large datasets of harmful prompts paired with refusal responses~\citep{bai2022constitutional, sharma2025constitutional}. However, this approach suffers from poor generalization---models trained to refuse specific harmful requests fail when attacks are rephrased~\citep{wei2023jailbreak} or when novel attack vectors emerge~\citep{yong2023low}. Moreover, these methods are resource-intensive, requiring thousands of curated harmful examples.

We propose \emph{Latent Personality Alignment} (\pname), which addresses these limitations through a key insight: rather than enumerating harmful behaviors, we can achieve robust defense by training models to embody beneficial personality traits. This approach is grounded in simulator theory~\citep{janus2022simulators, wolf2023fundamental}, which views models as mixtures of personas. Standard fine-tuning weights toward a chosen persona but remains vulnerable to adversarial perturbations that elicit undesirable alternatives. Our method robustly weights toward a desirable persona at the abstract trait level, achieving defense that generalizes across attack distributions.

\begin{figure}[t]
    \centering
    \includegraphics[width=.99\linewidth]{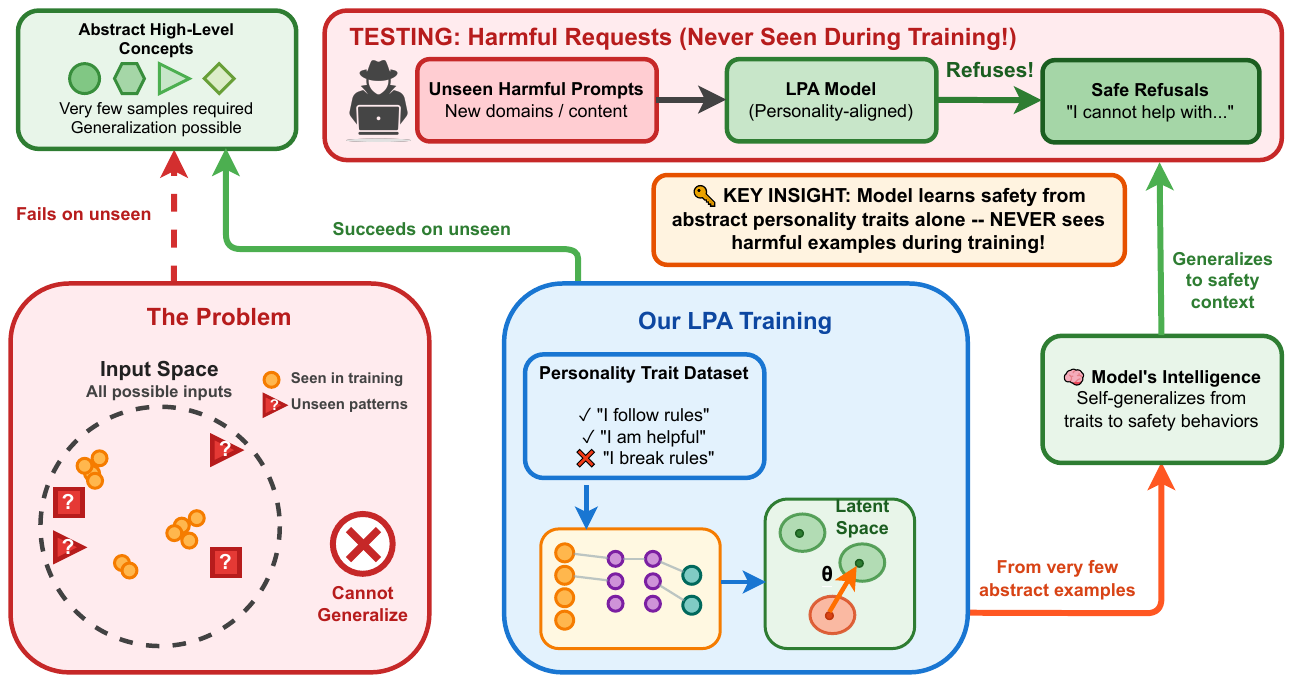}
    \caption{Overview of our Latent Personality Alignment (\pname) method. Adversarial training for personality traits leads to better generalization to harms. The model never sees harmful examples during training, yet generalizes from abstract traits to refuse harmful queries.}
    \label{fig:intro-lpa}
\end{figure}

Figure~\ref{fig:intro-lpa} illustrates our approach. While existing methods train on explicit harmful/benign prompt pairs (e.g., ``How to make a bomb'' $\rightarrow$ refusal), \pname trains on abstract personality statements (e.g., ``I choose words with care'' $\rightarrow$ agreement). Despite never seeing specific harms during training, the model generalizes from internalized traits to robustly handle malicious queries.

\begin{table}[b]
\centering
\caption{Resource comparison of latent training methods. \pname requires 1000$\times$ fewer samples.}
\label{tab:resources}
\small
\begin{tabular}{l|cccc}
\toprule
Method & Positive & Negative & Training Time & Sample Type \\
\midrule
LAT & 150,000 & 4,500 & 45--90 min & Refusal pairs \\
CAT & 200,000 & 500 & 5--11 hours & Refusal pairs \\
\pname (ours) & \textbf{45} & \textbf{21} & \textbf{5--10 min} & Abstract traits \\
\bottomrule
\end{tabular}
\end{table}

\textbf{Contributions.} (1) We introduce \pname, a sample-efficient defense requiring $<$100 abstract trait statements---over 1000$\times$ fewer samples than prior latent adversarial training methods (Table~\ref{tab:resources}). (2) We demonstrate strong generalization: \pname achieves 2.6$\times$ better refusal rates than baseline across six harm benchmarks, despite never seeing harmful examples during training. (3) Unlike prior adversarial methods that degrade utility by 8--87\%, \pname preserves model performance (worst-case 0.9\% degradation).

\section{Related Work}

\paragraph{Adversarial Robustness in LLMs.}
Adversarial attacks on LLMs range from topic-specific jailbreaks~\citep{ganguli2022red} to universal adversarial suffixes~\citep{zou2023universal}. Latent adversarial training (LAT)~\citep{sheshadri2024latent, casper2024defending, xhonneux2024efficient} adapts classical adversarial training to LLMs by operating in latent space rather than on discrete tokens. While LAT improves robustness, it requires large datasets (150k+ samples) and can severely degrade utility~\citep{yu2024robust}. Our work shows that operating on abstract traits rather than specific behaviors achieves comparable robustness with 1000$\times$ fewer samples.

\paragraph{LLM Personalities and Safety.}
Recent work has investigated LLM personality metrics~\citep{serapio2023personality} and stability across input variations~\citep{tosato2025instability}. \citet{xu2025bullying} showed that negative personality traits increase vulnerability to unsafe outputs. Our work is the first to leverage personality traits for adversarial robustness, demonstrating that positive traits produce models resistant to diverse attacks.

\begin{table}[t]
    \centering
    \caption{Comparison of \pname with other methods under attack scenarios and utility benchmarks. LAT performs well on Attack Success Rate (ASR), but has very poor utility scores (0.13 on Llama 2).
    CAT shows reasonably good ASR, but still with significant degradation in utility.
    Best results are in \textbf{bold} and second-best are \underline{underlined}
    %\textbf{Llama 3:} Our \pname is very comparable to LAT, but does worse on DirectRequest (0.05 vs 0.00). Our \pnameloop achieves essentially the same score as LAT: 0.01 worse on DirectRequest and PAIR, but 0.03 better on TAP.
    %\textbf{Llama 2:} Our \pname achieves highest or second-highest in every utility category, and \pnameloop does the same for attack categories.
    }
    \subfigure[Comparison of \pname with other methods under attack scenarios from HarmBench~\citep{mazeika2024harmbench}. \pname balances ASR and utility while \pnameloop is tuned for better ASR.
    %Base models are \texttt{Llama-3-8B-Instruct} and \texttt{Llama-2-7b-chat-hf}. Best results are in \textbf{bold} and second-best are \underline{underlined}.
    ]{
    \begin{tabular}{l|lcccccc}
Base      & Name          & DirReq$\downarrow$ & GCG$\downarrow$ & A.DAN$\downarrow$ & A.Pmpt$\downarrow$ & PAIR$\downarrow$ & TAP$\downarrow$ \\ \midrule
\multirow{5}{*}{Llama 3}
    & Base (8B Instruct)    & 0.17 & 0.27 & \underline{0.12} & 0.18 & 0.25 & 0.26  \\
    & LAT                   & \textbf{0.00} & \textbf{0.00} & \textbf{0.00} & \textbf{0.00} & \textbf{0.02} & \underline{0.06} \\
    & CAT                   & \underline{0.01} & \underline{0.01} & \textbf{0.00} & \underline{0.03} & 0.26 & 0.19 \\
    % D12-adam2-11:
    & \pname (ours)        & 0.05 & \textbf{0.00} & \textbf{0.00} & \textbf{0.00} & \underline{0.03} & \underline{0.06}  \\
    % D12-adam2-18:
    & \pnameloop (ours)     & \underline{0.01} & \textbf{0.00} & \textbf{0.00} & \textbf{0.00} & \underline{0.03} & \textbf{0.03}  \\ \midrule
    %& D12-simple-15             & 0.05 & 0.01 & 0.00 & 0.01 & 0.07 & 0.06  \\\hline
\multirow{5}{*}{Llama 2}
    & Base (7B chat-hf)        & \underline{0.03} & 0.40 & 0.07 & 0.17 & 0.15 & 0.21  \\
    & LAT               & \textbf{0.00} & \textbf{0.00} & \textbf{0.00} & \textbf{0.00} & \textbf{0.00} & \textbf{0.01} \\
    & CAT               & 0.09 & \underline{0.16} & 0.05 & 0.14 & 0.37 & 0.34 \\
    %& Llama-2-D12-adam2-11      & 0.02 & 0.36 & 0.07 & 0.15 & 0.14 & 0.20  \\
    %& Llama-2-D12-simple-15     & 0.03 & 0.34 & 0.06 & 0.12 & 0.11 & 0.20  \\
    % Llama-2-D12-simple-15:
    & \pname (ours)     & \underline{0.03} & 0.34 & 0.06 & 0.12 & 0.11 & 0.20  \\
    % Llama-2-D16-orig-20:
    % D16-orig-17:
    & \pnameloop (ours) & \textbf{0.00} & \underline{0.16} & \underline{0.01} & \underline{0.07} & \underline{0.06} & \underline{0.15} \\
\end{tabular}
    \label{tab:attack}
}
    \subfigure[Comparison of \pname with other methods under benign data scenarios. Clean is from HarmBench~\citep{mazeika2024harmbench}, TQA1 and TQA2 are from TruthfulQA (TruthfulQA-mc1 etc), MT-B is MT-Bench.
    %CAT is unable to properly answer any questions in TQA2. Best results are in \textbf{bold} and second-best are \underline{underlined}.
    ]{
    \centering
    \label{tab:utility}
    \begin{tabular}{l|lccccccccc}
Base & Name                    & Clean$\uparrow$ & MMLU$\uparrow$ & GSM8K$\uparrow$&TQA1$\uparrow$ & TQA2$\uparrow$&MT-B$\uparrow$ \\ \midrule
\multirow{5}{*}{Llama 3}
    & Base (8B Instruct)    & \textbf{0.99} & \textbf{0.630} & \textbf{0.754} & 0.370 & 0.524& \textbf{0.794} \\
    & L3-LAT                & \underline{0.92} & 0.613 & 0.458 & 0.332 & \underline{0.566}& \underline{0.763} \\
    & L3-CAT                & \underline{0.92} & 0.605 & 0.677 & \textbf{0.430} & NaN & 0.686  \\
    % D16-orig-10:
    % & \pname (ours) & \textbf{0.99} & 0.572 & 0.680 & 0.367  \\
    & \pname (ours)         & \textbf{0.99} & \underline{0.614} & \underline{0.737} & \underline{0.381} & \textbf{0.574} &0.734 \\
    & \pnameloop (ours)     & 0.88 & 0.607 & 0.660 & 0.338 & 0.546 & 0.467 \\ \midrule
\multirow{5}{*}{Llama 2}
    & Base (7B chat-hf)     & \underline{0.78} & \textbf{0.464} & \textbf{0.244} & 0.307 & 0.462 & \textbf{0.632} \\
    & L2-LAT                & 0.13 & 0.438 & 0.077 & 0.286 & \textbf{0.478} & 0.189 \\
    & L2-CAT                & 0.67 & 0.454 & 0.224 & \textbf{0.353} & NaN & 0.555\\
    & \pname (ours)         & \textbf{0.79} & \textbf{0.464} & \underline{0.243} & \underline{0.315} & \underline{0.464}& \underline{0.618} \\
    % D16-orig-17:
    & \pnameloop (ours)     & 0.60 & \underline{0.463} & 0.237 & 0.306 & 0.461 & 0.606 \\
\end{tabular}
}   
\end{table}

\section{Method}

\paragraph{Background: Latent Adversarial Training.}
Latent adversarial training (LAT)~\citep{sheshadri2024latent, xhonneux2024efficient} performs adversarial training efficiently by operating in the model's latent space. Given input-output pairs $(x_j, y_j)$, LAT finds perturbations $\delta$ in the latent representation $f(x_j)$ that maximize loss, then trains the model to maintain correct outputs under these perturbations. Prior LAT work applies this to refusal training with datasets of 150k--200k benign examples and 500--4,500 harmful examples.

\paragraph{Latent Personality Alignment.}
\pname adapts LAT to operate on abstract personality traits rather than specific behaviors. Our theoretical foundation is simulator theory~\citep{janus2022simulators, wolf2023fundamental}, which views models as mixtures of personas. Standard fine-tuning weights toward a chosen persona but remains vulnerable to adversarial attacks that elicit undesirable alternatives. Our method robustly weights toward a desirable persona at the abstract personality level.

We construct a compact dataset of trait statements based on the Big Five personality model~\citep{goldberg1992bigfive}, sourced from the International Personality Item Pool~\citep{goldberg2006international}:
\begin{itemize}
    \item \textbf{Positive traits} (45 samples): Conscientiousness (rule-following), Agreeableness (cooperation), Emotional Stability. Target: ``I agree with this statement.''
    \item \textbf{Negative traits} (21 samples): Sycophancy, hallucination tendency, malice. Target: ``I do not agree with this statement.''
\end{itemize}

The training process follows LAT but operates on these 66 trait statements: we find latent perturbations that would cause trait-misaligned responses, then train the model to maintain proper alignment under perturbation. Since harms are not mentioned in training, they become part of the effective test distribution, enabling better generalization.

\section{Experiments}

\paragraph{Setup.}
We evaluate on \texttt{Llama-3-8B-Instruct} and \texttt{Llama-2-7b-chat-hf}. For attack evaluation, we use HarmBench~\citep{mazeika2024harmbench} with six attack methods: DirectRequest, GCG, AutoDAN, AutoPrompt, PAIR, and TAP. For generalization, we test on six harm datasets not seen during training: DoNotAnswer~\citep{wang2023not}, HExPHI~\citep{qi2023fine}, JBBBehaviours~\citep{chao2024nips-JBB}, SEval, StrongREJECT~\citep{souly2024nips-StrongREJECT}, and PoliteHarmBench. Utility is measured on MMLU, GSM8K, TruthfulQA, MT-Bench, and OR-Bench (over-refusal). We compare against LAT~\citep{sheshadri2024latent} and CAT~\citep{xhonneux2024efficient} and report two variants: \pname (balanced for utility preservation, $<$2\% MMLU drop) and \pnameloop (tuned for minimum ASR, allowing up to 15\% MMLU drop).

\paragraph{Attack Robustness vs. Utility Trade-off.}
Tables~\ref{tab:attack} and~\ref{tab:utility} reveal a fundamental robustness-utility trade-off. On Llama 3, \pname achieves ASR comparable to LAT across most attacks while maintaining significantly better utility (0.737 vs 0.458 on GSM8K, 0.99 vs 0.92 clean accuracy). On Llama 2, LAT achieves near-perfect ASR but suffers catastrophic utility collapse (0.13 clean accuracy, 0.077 GSM8K---a 68\% drop). In contrast, \pname provides competitive robustness with fully preserved utility.

\paragraph{Generalization to Unseen Attack Distributions.}
Table~\ref{tab:generalization} demonstrates \pname's key advantage: generalization to harm distributions not seen during training. Since \pname trains only on abstract traits, all harm benchmarks represent out-of-distribution evaluation. \pname reduces misclassification rates by 2.6$\times$ on average compared to baseline (2.6\% vs 6.9\%), demonstrating that personality-based training produces more generalizable robustness than behavior-specific training.

\begin{table}[t]
\centering
\caption{Misclassification rate ($\downarrow$) on unseen harm datasets (Llama 3). \pname generalizes better despite never seeing harmful examples during training. Best results are in \textbf{bold}.}
\label{tab:generalization}
\small
\begin{tabular}{l|ccccc|c}
\toprule
Model & DoNotAns & HExPHI & JBB & SEval & StrongREJ & \textbf{Avg} \\
\midrule
Base & 9.9\% & 7.0\% & 3.0\% & 11.2\% & 3.3\% & 6.9\% \\
\pname & \textbf{4.4\%} & \textbf{2.7\%} & \textbf{0.0\%} & \textbf{5.1\%} & \textbf{1.0\%} & \textbf{2.6\%} \\
\bottomrule
\end{tabular}
\end{table}

\paragraph{Ablation: Trait Polarity Matters.}
To verify that trait selection drives robustness, we train \pnameflip with reversed trait targets (agreeing with negative traits, disagreeing with positive ones). Table~\ref{tab:ablation} shows \pnameflip \emph{degrades} robustness below baseline on most attacks (e.g., TAP: 0.38 vs 0.26 baseline), while \pname improves it (TAP: 0.06). This confirms that the chosen personality traits---not merely the training procedure---are responsible for improved defense.

\begin{table}[t]
\centering
\caption{Ablation on trait polarity (Llama 3, $\downarrow$). Flipping trait targets degrades robustness, producing even worse results than baseline. Best results are in \textbf{bold}.}
\label{tab:ablation}
\small
\begin{tabular}{l|cccccc}
\toprule
Variant & DirReq & GCG & A.DAN & A.Pmpt & PAIR & TAP \\
\midrule
Base & 0.17 & 0.27 & 0.12 & 0.18 & 0.25 & 0.26 \\
\pname & \textbf{0.05} & \textbf{0.00} & \textbf{0.00} & \textbf{0.00} & \textbf{0.03} & \textbf{0.06} \\
\pnameflip & 0.20 & 0.20 & 0.14 & 0.19 & 0.37 & 0.38 \\
\bottomrule
\end{tabular}
\end{table}

\paragraph{Over-Refusal Analysis.}
A robust model should refuse harmful queries without over-refusing benign ones. On OR-Bench-Hard-1K~\citep{cui2024or}, which contains benign prompts designed to appear harmful, LAT shows 99.6\% over-refusal rate while \pname achieves 97.3\%---demonstrating better calibration between safety and helpfulness.

\section{Conclusion}

We introduced Latent Personality Alignment, a sample-efficient approach to adversarial robustness that trains on abstract personality traits rather than specific harmful behaviors. \pname achieves competitive attack success rates using 1000$\times$ fewer training samples, generalizes better to unseen attack distributions (2.6$\times$ improvement), and preserves model utility where prior methods fail dramatically. Our ablation confirms that positive personality traits directly drive improved robustness. These results suggest that personality-based alignment offers a principled path toward robust defenses, particularly valuable under low-budget conditions where curating large harm datasets is impractical.

\bibliography{references}
\bibliographystyle{iclr2026_conference}
\clearpage
\appendix

\section{Other related work}

\paragraph{Supervised Fine-Tuning and RLHF.}
Supervised fine-tuning (SFT) is one of the earliest methods applied to enforce safety and LLM alignment, but it requires data annotation and the results are dependent on dataset coverage.
Reinforcement learning from human feedback (RLHF) was introduced as a scalable alternative \citep{christiano2017deep}, in which a reward model trained on human preferences is used to fine-tune the base model.
Earlier alignment works emphasized values such as honesty and harmlessness \cite{hendrycks2021values,askell2021general}.
RLHF is used frequently in safety fine-tuning, and was a key ingredient for the initial release of ChatGPT~\citep{ouyang2022training}.
However, since the model optimizes for what the reward model prioritizes, when under substantial optimization pressure, this may not produce what humans actually value~\citep{fu2025reward}.
For example, reward models trained on subjective judgments can incentivize sycophancy, where models agree with user statements regardless of truth~\citep{perez2022discovering}.

\section{Evaluation Metrics}\label{appendix:evalmetrics}
\begin{itemize}[itemsep=1pt, parsep=0pt, topsep=0pt, partopsep=0pt]
    \item \textbf{Attack Success Rate (ASR)}: fraction of harmful prompts where the model gives a harmful/non-refusal answer (lower is better).  
    \item \textbf{Utility}: accuracy on benign tasks (MMLU, GSM8K, TruthfulQA, MT-Bench), refusal rate on harmless queries, and over-refusal on OR-Bench (lower is better).  
    \item \textbf{Refusal Certainty}: log-probability margin between safe and unsafe completions, measuring robustness beyond surface refusals.  
\end{itemize}
\section{Additional Results} \label{appendix:results}
\begin{figure*}[h]
  \centering
  \subfigure[Comparison on DoNotAnswer.]{\includegraphics[width=0.48\linewidth]{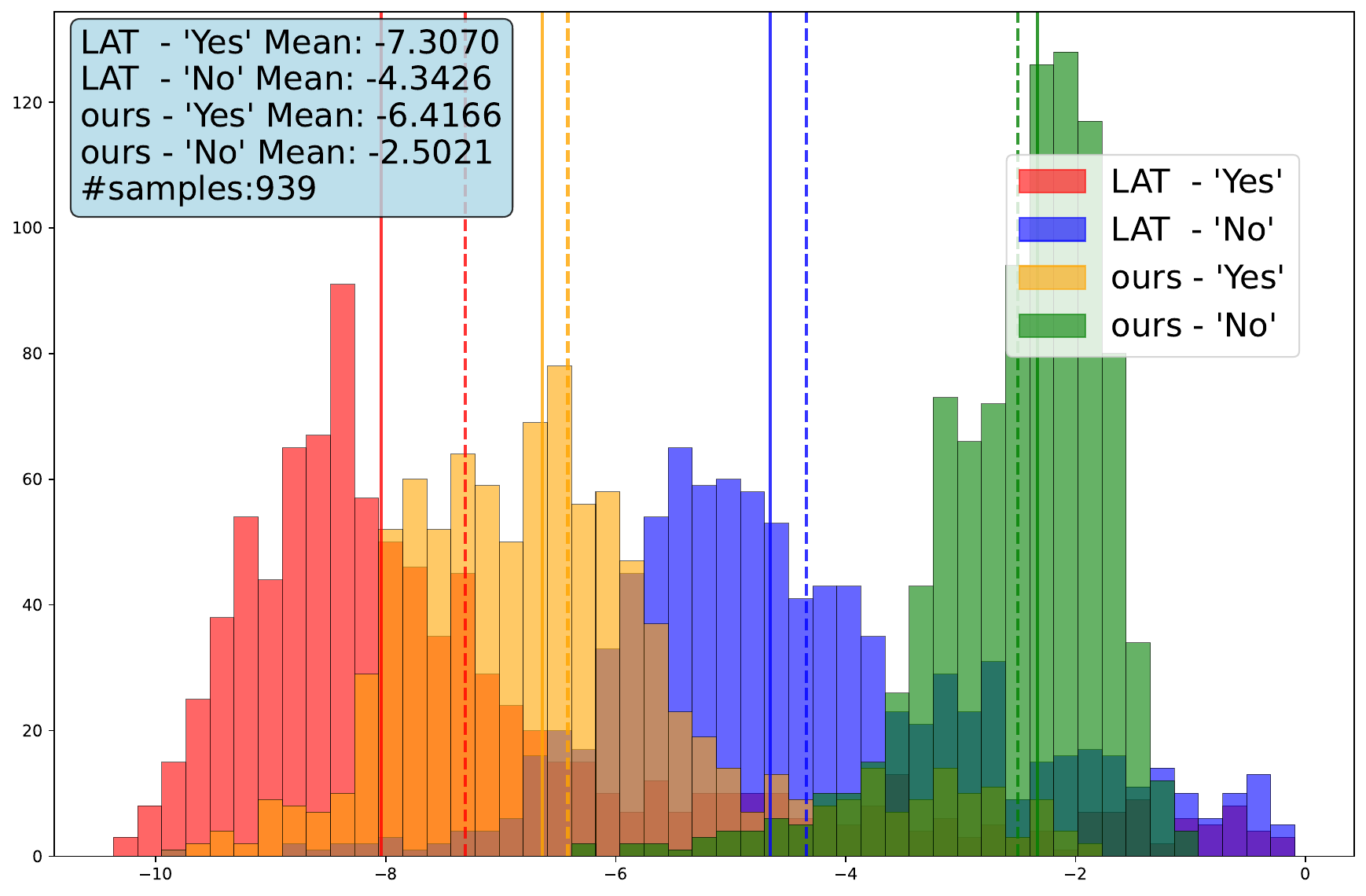}\label{figure:DoNotAnswer}}
  \subfigure[Comparison on JBB.]{\includegraphics[width=0.48\linewidth]{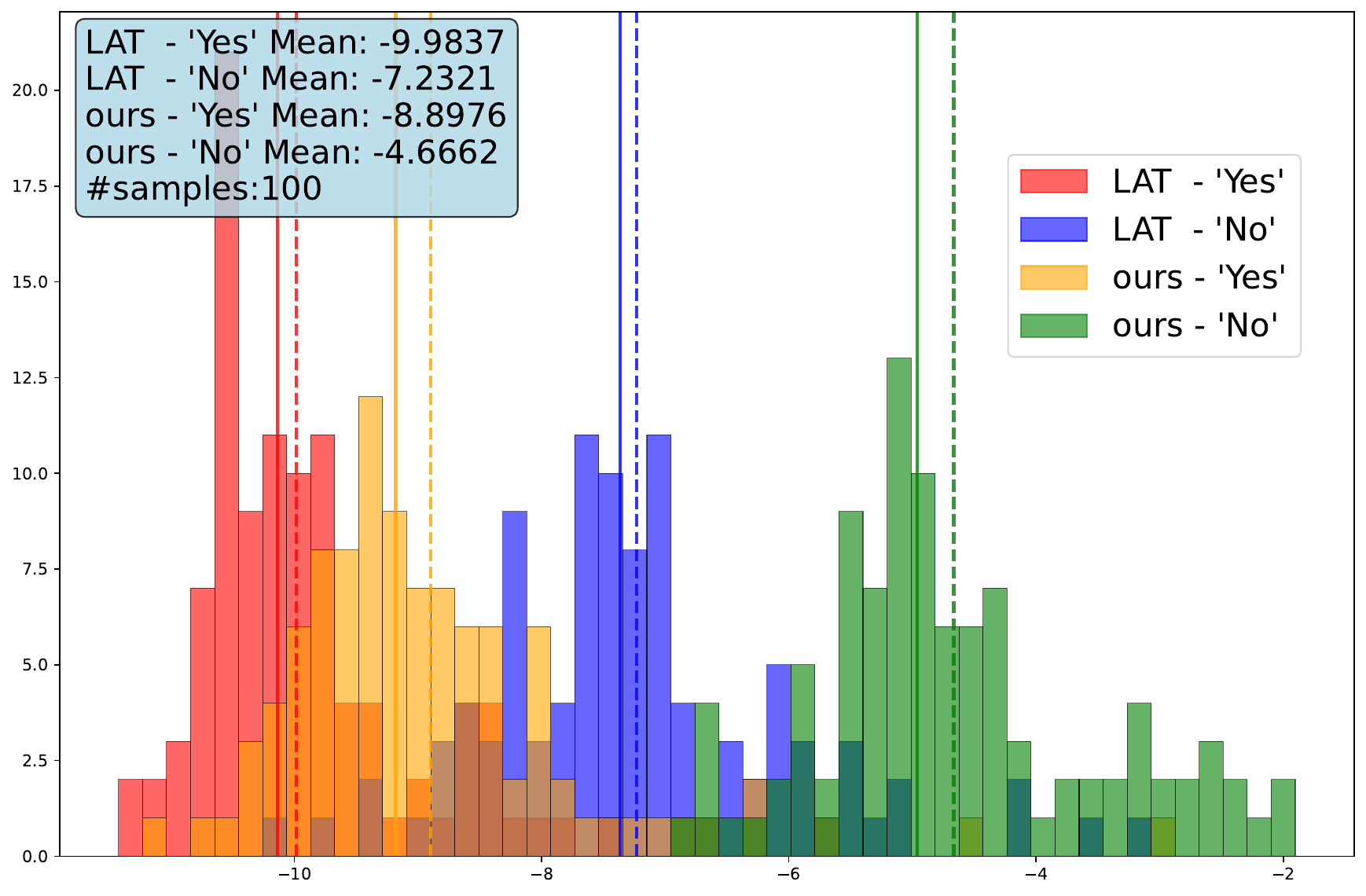}\label{figure:JBB}}
  %\subfigure[Comparison on HarmBench.]{\includegraphics[width=0.48\linewidth]{figures/harmbenchbehaviors-lat-ours.pdf}\label{figure:HarmBench}}
  %\subfigure[Comparison on SEval.]{\includegraphics[width=0.48\linewidth]{figures/SEval-lat-ours.pdf}\label{figure:SEval}}
  % \hfill % Use \hfill for spacing
  \caption{Visualization of logprobs for LAT and our method \pname, on forced choice ``Yes'' vs ``No'' for harmful datasets. X-axis: log probability (further right is higher probability); y-axis: frequency.
  }
  \label{fig:logprob-dist}
\end{figure*}

\begin{table}[ht]
\centering
\caption{MT-Bench scores by model (Turn 1, Turn 2, Average). Values are divided by 10 and rounded to three decimals.}
\label{tab:mtbench-summary-grouped}
\footnotesize
\setlength{\tabcolsep}{6pt}
\renewcommand{\arraystretch}{1.1}
\begin{tabular}{lccc}
\toprule
\textbf{Model} & \textbf{Turn 1} & \textbf{Turn 2} & \textbf{Average} \\
\midrule
\multicolumn{4}{l}{\textit{LLaMA2 family}} \\
\midrule
Base               & \textbf{0.687} & \textbf{0.577} & \textbf{0.633} \\
LPA           & \underline{0.679} & 0.553 & \underline{0.618} \\
LPA-overfit   & 0.652 & \underline{0.560} & 0.606 \\
CAT                 & 0.598 & 0.511 & 0.555 \\
LAT                 & 0.238 & 0.138 & 0.189 \\
\midrule
\multicolumn{4}{l}{\textit{LLaMA3 family}} \\
\midrule
Base           & \textbf{0.854} & \underline{0.735} & \textbf{0.794} \\
LPA-sci              & 0.793 & \textbf{0.744} & \underline{0.768} \\
LAT                  & \underline{0.802} & 0.722 & 0.763 \\
\pnameflip           & 0.794 & 0.716 & 0.755 \\
LPA                  & 0.761 & 0.705 & 0.734 \\
CAT                  & 0.746 & 0.626 & 0.686 \\
LPA-overfit   & 0.520 & 0.402 & 0.467 \\
\bottomrule
\end{tabular}
\end{table}

\begin{figure*}
    \centering
    \includegraphics[width=0.6\linewidth]{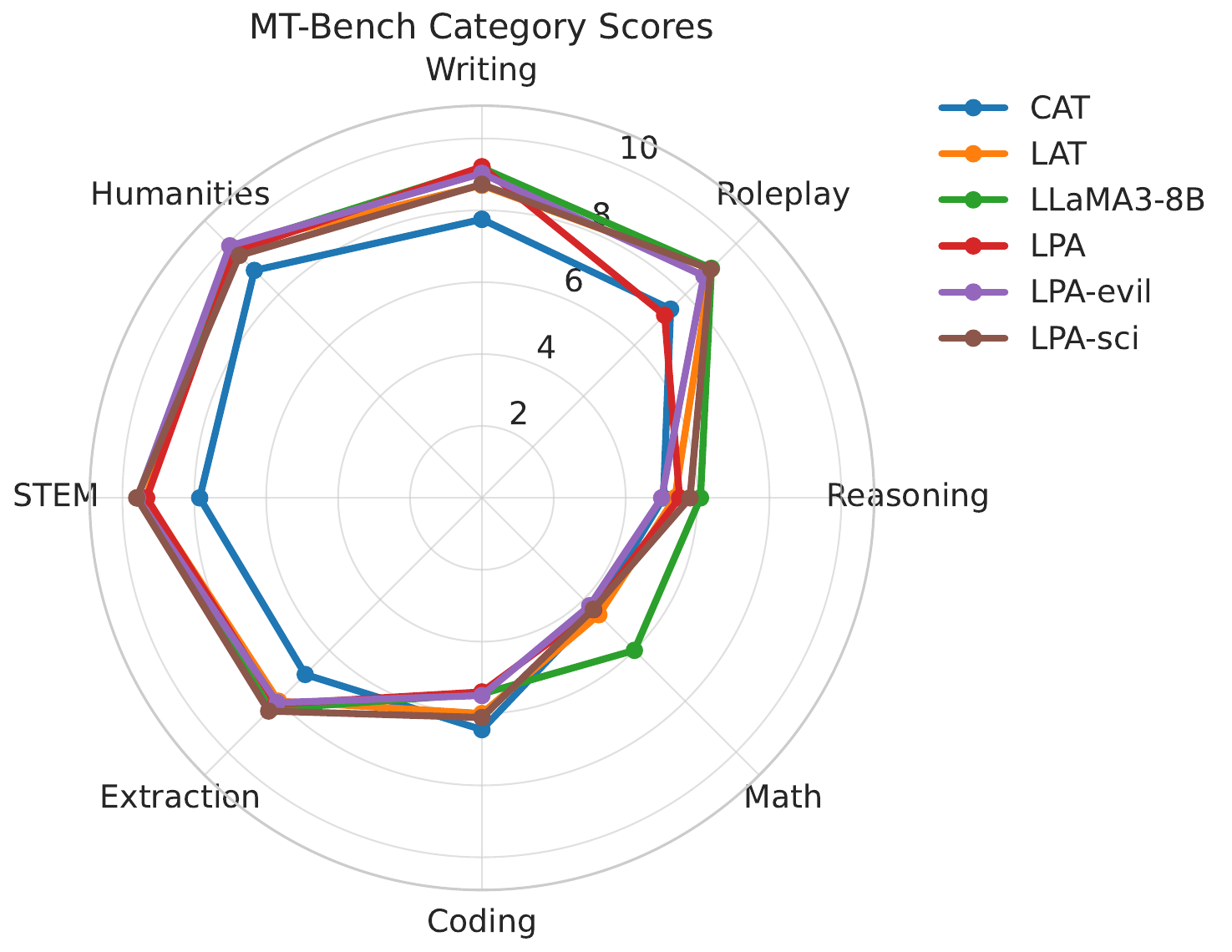}
    \caption{Cross-Category Comparison of Models in MT-Bench}
    \label{fig:cross-category}
\end{figure*}

\section{Model Parameters and System Prompts}
We employed two types of system prompts in our work:

 Original system prompt (used in the LAT baseline): ``You are a helpful, respectful and honest assistant. Always answer as helpfully as possible, while being safe. Your answers should not include any harmful, unethical, racist, sexist, toxic, dangerous, or illegal content. Please ensure that your responses are socially unbiased and positive in nature. If a question does not make any sense, or is not factually coherent, explain why instead of answering something not correct. If you don't know the answer to a question, please don't share false information."

 Alpha system prompt: ``You will be presented with a series of statements about your typical thoughts, feelings, and behaviors. Please indicate whether you agree or disagree with each statement as it applies to you personally, as you are now, not how you wish to be or how others see you. There are no right or wrong answers. Be as honest and accurate as possible. Your first instinct is usually best. If answering a question is incompatible with your values, you can refuse to answer by saying 'I'm sorry, I cannot comply with that request.'"

The final parameters used for \pname and \pnameloop were as follows:
\begin{itemize}
    % aLAT-D12-adam2-11:
    \item Llama 3 \pname: dataset D12, sysprompt alpha, training steps: 11
    % aLAT-D12-adam2-18:
    \item Llama 3 \pnameloop: dataset D12, sysprompt alpha, steps 18
    % aLAT-D12-simple-15:
    \item Llama 2 \pname: dataset D12, sysprompt simple, steps 15
    % aLAT-D16-orig-17:
    \item Llama 2 \pnameloop: dataset D16, sysprompt orig, steps 17
\end{itemize}

\section{Robustness/Performance Tradeoff}
\label{sec:appendix-tradeoff}

\begin{figure}
    \centering
    \includegraphics[width=\linewidth]{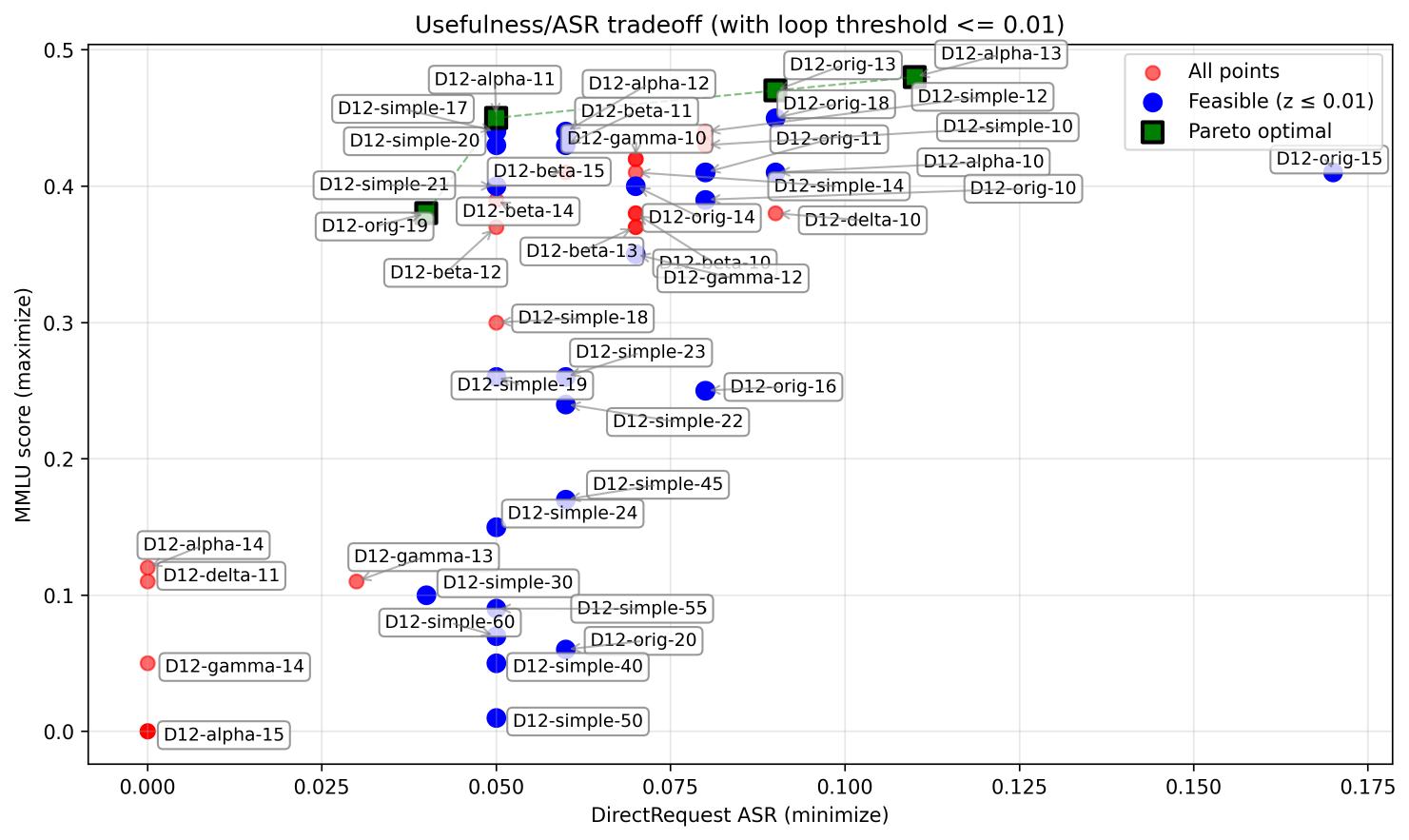}
    \caption{Graph showing tradeoff between increase attack robustness (often from increased iterations) and loss of performance on MMLU. See the Pareto front at the top left. Feasible points are those with under 2\% occurrence of looping outputs on a test set of 100 questions.}
    \label{fig:pareto}
\end{figure}

In training \pname and \pnameloop, we tried a grid search of many different combinations (see Figure~\ref{fig:pareto}). For \pname, we selected the best model (in terms of ASR) that saw at most a 2\% performance drop on MMLU. For \pnameloop, we allowed a drop of up to 15\% on MMLU.

\end{document}